\documentclass[10pt]{article}
\usepackage{amsmath,epsfig}
\usepackage{amssymb}
\usepackage{graphicx}
\usepackage{color}
\usepackage{alphalph}
\usepackage{url}
\usepackage{colortbl}
\usepackage{authblk}
\usepackage{epstopdf}
\usepackage{slashbox}
\usepackage{subcaption}
\usepackage{enumitem}
\usepackage{hyperref}
\usepackage{tabularx}
\usepackage{multirow}
\usepackage{booktabs}
\usepackage[export]{adjustbox}
\usepackage[justification=centering]{caption}
\usepackage{xcolor}
\usepackage{tikz}
\usepackage{pgfplots}
\pgfplotsset{compat=1.18}
\usepackage{authblk}
{\itshape\small}

\usepackage{titlesec}
\titleformat{\paragraph}
{\normalfont\normalsize\bfseries}{\theparagraph}{1em}{}
\titlespacing*{\paragraph}
{0pt}{3.25ex plus 1ex minus .2ex}{1.5ex plus .2ex}
\newcolumntype{M}[1]{>{\centering\arraybackslash}m{#1}}

\begin{document}

\newcommand\myworries[1]{\textcolor{red}{#1}}

\title{MLLM Frame Subset Ensembling for Audio-Visual Video QA and MLLM-based Reranking for Ad-hoc Video Search in TRECVID 2025}

\author[1,2]{Andreas Goulas}
\author[1]{Damianos Galanopoulos} 
\author[2]{Ioannis Patras} 
\author[1]{Vasileios Mezaris}
\affil[1]{Information Technologies Institute (ITI),
Centre of Research and Technology Hellas (CERTH),
Thessaloniki, Greece}

\affil[2]{Queen Mary University of London, London, UK}
\affil[ ]{\{agoulas, dgalanop, bmezaris\}@iti.gr, i.patras@qmul.ac.uk}
\date{}
\maketitle

\section*{\centering Abstract}
This paper presents the overview of the runs related to the Ad-hoc Video Search (AVS) and Video Question Answering (VQA) tracks of TRECVID 2025 on behalf of the CERTH-ITI team. For the AVS track, we introduce a two-stage framework built on foundation models. In the first stage, multiple vision–language models (VLMs) encode both the input query, augmented through LLM-generated rephrasings, and the candidate video shots, producing weighted similarity scores for initial retrieval. In the second stage, we utilize a Multimodal-LLM(MLLM)-based reranking module that evaluates the semantic alignment between each shot among the top-N highest-ranked ones and the original query, generating updated relevance scores for reordering these shots. This MLLM-driven reranking significantly improves contextual matching and produces more accurate final rankings without requiring any model training. 
Regarding the VQA track, we fine-tune an audio-visual MLLM model on the provided TRECVID training dataset and we implement an inference-time scaling technique to enhance the multimodal understanding capabilities of the MLLM. For the open-ended Answer Generation (AG) task, we aggregate multiple model responses per question via a majority vote. The responses are generated with greedy sampling from different random frame subsets of the video and they are ranked based on the number of votes.
For the Multiple-Choice (MC) task, instead of voting, we use mean pooling on the logits assigned by the fine-tuned model to each candidate response.
Through the combination of fine-tuning and frame subset ensembling we achieve the highest score across 3 metrics in the VQA AG task and the second highest in the VQA MC task.

\section{Introduction}
CERTH-ITI has been participating in TRECVID \cite{trecEval}, either in collaboration with other organizations under the umbrella of a specific EU-funded project, or as a stand-alone organization, since 2004. In this paper, we present the CERTH-ITI approaches for the Multimedia Tracks of TREC 2025~\cite{2025trec}, specifically the Ad-hoc Video Search (AVS)~\cite{2025trecavs} and the Video Question Answering (VQA)~\cite{2025trecvqa} tracks. The goal of an AVS method is to retrieve relevant video shots using free-text queries. On the other hand, a VQA method focuses on answering questions related to a specific video, either in an open-ended format or through multiple-choice options. The recent advances in foundation models enable more robust and semantically aligned representations across text and visual domains. This year, we built upon such foundation models to create reliable and well-performing methods on both tracks. For the AVS track, our approach focuses on utilizing foundation models to improve text–video retrieval performance. Building upon our previous submissions~\cite{trecvid2023,trecvid2024}, we adopt a fully training-free pipeline that integrates multiple pre-trained Vision–Language Models (VLMs) for video shot retrieval, and Large Language Models (LLMs) for query augmentation and contextual re-ranking.
For the VQA track, we fine-tune a pre-trained audio-visual MLLM to address the two VQA tasks and, during inference, we scale the capabilities of the model by aggregating multiple predictions from the model, generated with greedy sampling. Specifically, for each generation the model sees a different random subset of the video frames.
The following sections will present the employed algorithms and the evaluation of our runs in the AVS and VQA tracks, respectively.

\section{Ad-hoc Video Search} \label{SS:AVS_SYSTEM}

The TRECVID 2025 \cite{2025trec} Ad-hoc Video Search (AVS) task aims to develop systems capable of retrieving a ranked list of 1,000 video shots for each free-form textual query, ordered from the most to the least relevant. Building on our previous participations, we address this task by leveraging a diverse set of pre-trained image-text models to enhance AVS performance. Moreover, the recent rise of multimodal large language models (MLLMs)~\cite{qwen2.5, bai2025qwen2} has expanded the possibilities for text-based video retrieval. Motivated by these developments, we propose a new two-stage framework that integrates multiple VLMs for image-based retrieval, an LLM  for query augmentation and an MLLM for contextual re-ranking.
In contrast to our earlier approaches~\cite{trecvid2024,trecvid2023}, which combined outputs of several cross-modal networks through a trainable neural aggregation module, this year’s system does not involve any training. Our internal experiments showed that additional training provided only marginal gains while requiring substantial computational resources. To balance performance and efficiency, we adopt a fixed weighting scheme. Furthermore, we incorporate two LLMs: one for augmenting the input queries and an MLLM to rerank the retrieved video shots by producing contextual relevance scores that assess how well each shot matches the query.

\begin{figure*}[t]
  \centering
  
   \includegraphics[width=0.95\linewidth]{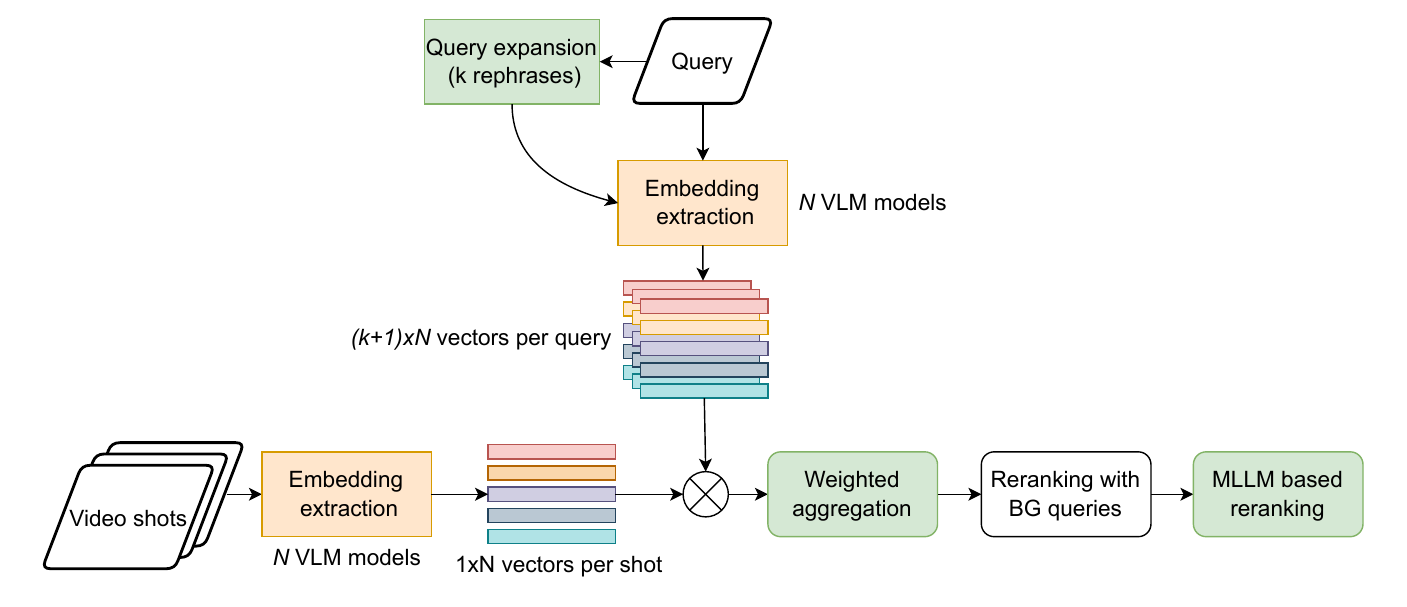}

   \caption{An overview of the proposed AVS method.}
   \label{fig:all}
\end{figure*}

\subsection{Approach}
In our AVS 2025 submission, we build upon and extend our AVS 2024 approach by introducing a refined two-stage retrieval–reranking framework. The complete AVS architecture is shown in Figure~\ref{fig:all}. In the retrieval stage, the system analyzes the user query, generates multiple query augmentations, extracts VLM-based embeddings, computes similarities between the (augmented) query and all candidate videos, and retrieves the most relevant video shots. In the reranking stage, we leverage an MLLM to reassess and refine these retrieval results by evaluating how well each retrieved video matches the original query.

In the retrieval stage, in order to enrich the semantic representation of the input query $q$, we generate multiple variations that reflect different levels of textual detail and help reveal implicit information. For this step, we employ \textit{LLaMA-3.2-1B-Instruct}~\cite{meta2024llama, grattafiori2024llama3herdmodels}, prompting it to rephrase the query (Table \ref{tab:prompts}) while strictly preserving all visual details, entities, and attributes. This process yields a set of augmented queries
$ Q = \{ q, q_1, \ldots, q_k \} $.

We use five VLM families—CLIP~\cite{clip_2021}, BLIP~\cite{li2022blip}, BLIP-2~\cite{li2023blip}, SLIP~\cite{mu2022slip}, and BEiT3~\cite{beit3}—along with their respective pretrained variants (listed in Table~\ref{tab:models_AVS}). All video shots in our dataset are processed using these models to extract their embeddings.
Each augmented query in $Q$ is then encoded using the same VLMs so that both queries and video shots lie in a shared embedding space. For every query–video pair, we compute cosine similarity, and aggregate the similarities within each VLM family using mean pooling to produce a single relevance score per family. Finally, we fuse the scores from all families using a weighted combination with fixed weights across VLM families. The final weights, i.e., $[10:5:5:1:20]$ for CLIP, BLIP, BLIP-2, SLIP, and BEiT3, respectively, were determined by internal experiments on the 2022 and 2023 queries.
Following our 2024 approach~\cite{trecvid2024}, where we applied a similarity normalization procedure to enhance query–video shot matching, we utilize a normalization step to improve the comparability of similarity scores across queries. Specifically, for each query–video pair, we extend their final similarity score by comparing the same video shot against a set of background queries. This produces a vector consisting of the similarity between the target query and the shot, along with the similarities between the shot and all background queries. The resulting vector is then L2-normalized, ensuring that the normalized similarity of the target query accounts for how strongly the shot matches generic or unrelated background queries. After applying this normalization, the revised query-video shot similarity is the final similarity used for the retrieval stage, and using these similarity scores we generate a ranked list of the top-N most relevant video shots. As background queries, we consider the 2022, 2023, and 2024 AVS queries.

The second stage of our framework focuses on refining the initial top-N video shots returned in the retrieval stage for each input query. For this purpose, we employ the \textit{Qwen2.5-VL-7B} model~\cite{bai2025qwen25vl}, an instruction-tuned MLLM capable of evaluating fine-grained semantic alignment between textual descriptions and visual content. Given a query and each of the top-N retrieved video shots, \textit{Qwen2.5-VL-7B} jointly processes the text and visual frames and produces a contextual relevance score indicating how well each shot matches the intended semantics of the query. The prompt that was used is shown in Table \ref{tab:prompts}. These scores are then used to revise the original query-video shot similarities, yielding a final ranking that more accurately reflects the query-specific visual relevance.

\begin{table}
    \centering
        \caption{The prompts that were used for query augmentation and for the video shot reranking. }
    \begin{tabular}{>{\centering\arraybackslash}p{0.2\linewidth}|>{\centering\arraybackslash}p{0.7\linewidth}}\toprule
         LLM&  Prompts\\\midrule
             \textit{LLaMA-3.2-1B-Instruct }(query expansion)     &             You are given a short visual description used for video retrieval, such as  \textless \textit{query}\textgreater. Your task is to rephrase the sentence while preserving **all** visual details, entities, and attributes. Do not remove or replace any element. Only change the phrasing or order of words. 
 Provide twenty alternatives. Original: \textless \textit{query}\textgreater\\ \hline
         \textit{Qwen2.5-VL-7B }

(reranking)&  How relevant is this image to the following caption:  \textless \textit{query}\textgreater? 

Answer with a single number 1-10.\\ \bottomrule
    \end{tabular}

    \label{tab:prompts}
\end{table}

\begin{table}[ht]
\centering
\caption{The cross-modal network families and their pre-trained model variations utilized in our AVS 2025 participation.}
\label{tab:models_AVS}
\begin{tabular}{c|c||c|c}
\hline
\textbf{Network Family} & \textbf{Pre-trained Model} & \textbf{Network Family} & \textbf{Pre-trained Model} \\\hline
\multirow{4}{*}{BLIP} & base\_coco & \multirow{4}{*}{BEiT3} & base\_coco (retrieval) \\
 & base\_flickr &  & base\_flickr30k (retrieval) \\
 & large\_coco &  & large\_coco (retrieval) \\
 & large\_flickr &  & large\_flickr30k (retrieval) \\\hline
\multirow{3}{*}{BLIP-2} & itm\_coco & \multirow{8}{*}{CLIP} & RN101 \\
 & itm\_pretrain &  & RN50 \\
 & itm\_pretrain\_vitL &  & RN50x4 \\ \cline{1-2}
\multirow{5}{*}{SLIP} & base\_CC12M &  & RN50x16 \\
 & base\_CC3M &  & RN50x64 \\
 & base &  & ViT-B/16 \\
 & large &  & ViT-B/32 \\
 & small &  & ViT-L/14 \\\hline
\end{tabular}
\end{table}

\subsection{Submission}
The V3C2 \cite{rossetto2019v3c} dataset is utilized to evaluate the network's performance. Moreover, we examine the performance of our runs on the  V3C2 datasets for the queries of years 2022-2023 as well as the performance using the 2024 ground truth in the same query set of 2025. The evaluation measure we use is the mean extended inferred average precision (MxinfAP).

This year, we submitted four runs for the AVS 2025 main task. The submitted runs are briefly described below:

\begin{itemize}
\item ITI\_CERTH.25\_run\_1: Query expansion using $k=20$ automatically generated rephrasings of the original query; Reranking applied to the retrieved video shots up to a depth of $N=4,000$ candidates.
\item ITI\_CERTH.25\_run\_2: Similar to run\_1, using $N=2,000$ candidates in the reranking stage.
\item ITI\_CERTH.25\_run\_3: Similar to run\_1, using $N=1,000$ candidates in the reranking stage.
\item ITI\_CERTH.25\_run\_4: Base model;  Similar to ITI\_CERTH.25\_run\_1 without the reranking stage.

\end{itemize}

\subsection{Experimental Results}

\begin{table}[htb]
\caption{Mean Extended Inferred Average Precision (MXinfAP) for all submitted runs for the  2022, 2023, 2024 and 2025 fully automatic AVS tasks. Best results are \textbf{emphasized}.}\label{tab:results_AVS_main}
\begin{center}
{
\begin{tabular}{c|cccc}
\hline
\textbf{Run id:}        & \textbf{2022}     & \textbf{2023}& \textbf{2024}& \textbf{2025}\\ \hline
ITI\_CERTH.25 run\_1    &\textbf{0.332} &	\textbf{0.379}&	\textbf{0.416}&	\textbf{0.421}\\\hline

ITI\_CERTH.25 run\_2   &0.331&	0.378&	0.412&	0.41\\\hline
ITI\_CERTH.25 run\_3    &0.312&	0.357&	0.381&	0.369\\\hline
ITI\_CERTH.25 run\_4    &0.284	&0.330&	0.344&	0.337\\\hline
 
\end{tabular}}
\end{center}
\end{table}

\begin{figure}[htb]
\centering
\includegraphics[width=0.85\textwidth]{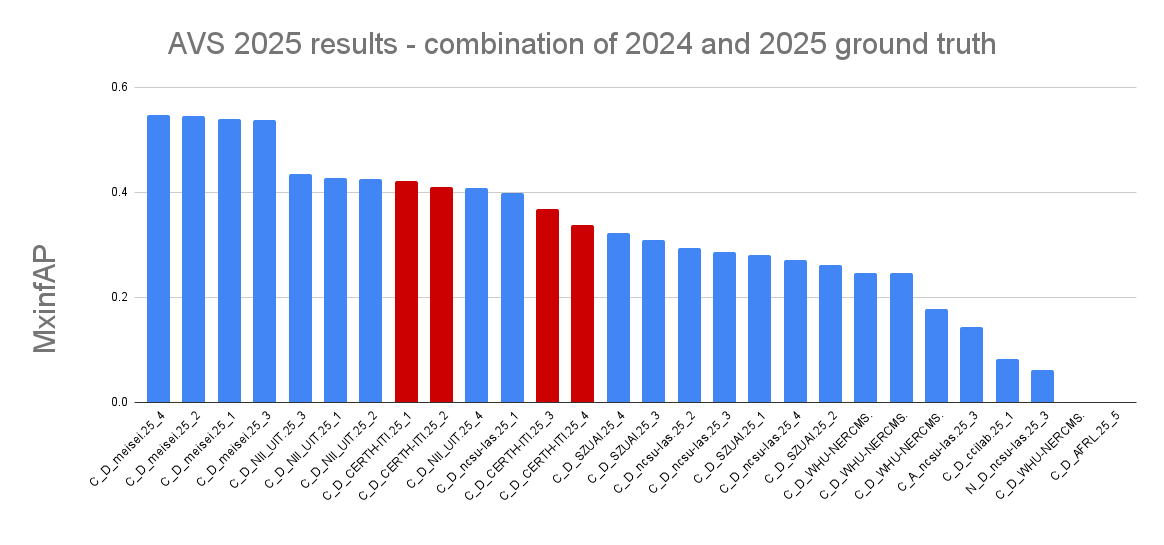}
\caption{AVS 2025 ranking of all submitted runs using the combined ground truth of 2024 and 2025 evaluations, in MXinfAP terms. Red bars indicate our submitted runs.}
\label{fig:result25}
\end{figure}

 \begin{figure}[htb]
\centering
\includegraphics[width=0.85\textwidth]{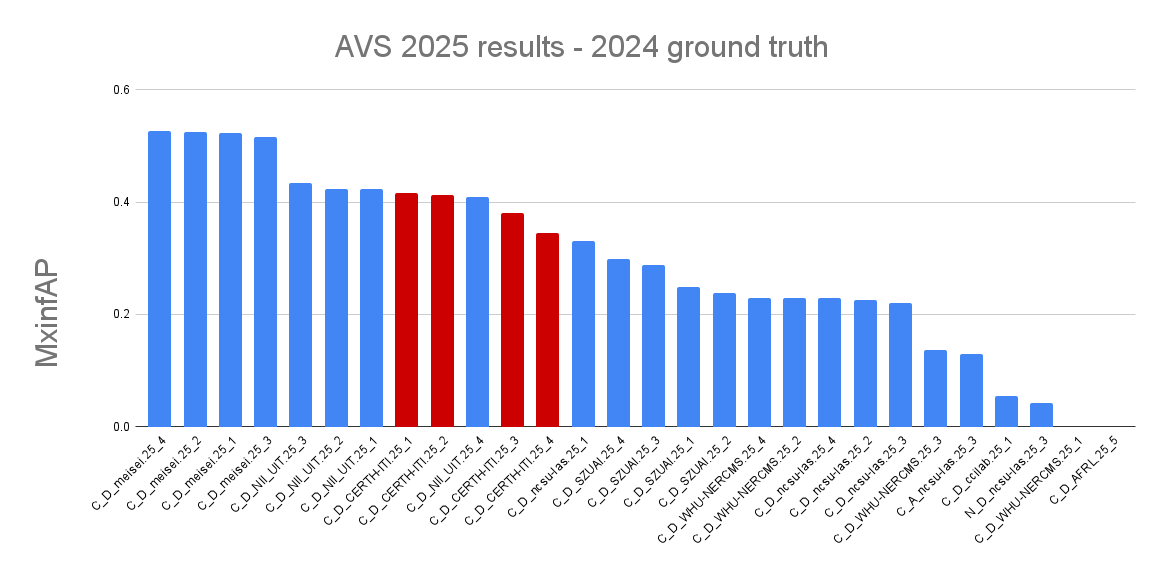}
\caption{AVS 2025 ranking of all submitted runs using only the 2024 ground truth, in MXinfAP terms. Red bars indicate our submitted runs.}
\label{fig:result24}
\end{figure}
 
 Table \ref{tab:results_AVS_main} presents the official results of our submissions for the main AVS 2025 task using the 2024 ground truth version and the updated 2025 one, as well as the results of our internal evaluation of AVS 2022 and 2023 queries. Overall, ITI\_CERTH.25\_run\_1 consistently achieves the highest MXinfAP scores in all four years, indicating that combining query expansion with reranking over a 4,000-shot depth provides the most effective configuration. Reducing the reranking depth to 2,000 shots (ITI\_CERTH.25\_run\_2) yields only a slight performance drop, indicating that the MLLM-based reranking remains effective even with fewer candidates. However, when the reranking depth is further reduced to 1,000 candidates (ITI\_CERTH.25\_run\_3), performance decreases more noticeably across all years, suggesting that a larger reranking pool is beneficial for capturing relevant shots missed in early retrieval. The base system without reranking (ITI\_CERTH.25\_run\_4) consistently produces the lowest scores, confirming the significant contribution of the MLLM reranking stage. The gains between ITI\_CERTH.25\_run\_4 and the top-performing CERTH runs highlight the importance of the contextual reranking in improving retrieval performance. Across all runs, the results are stable over the four evaluation years, showing that the proposed two-stage foundation-model-based pipeline generalizes well across different query sets.

Figure \ref{fig:result25} compares the performance of our submitted runs in the AVS 2025 main task against all submitted runs from the other participating teams, using the 2025 additionally-annotated ground truth data; Figure \ref{fig:result24} shows the same comparison when using just the 2024 ground truth.

\section{Video Question Answering}
\label{SS:VQA_SYSTEM}

\begin{figure}[htb]
\centering
\includegraphics[width=0.95\textwidth]{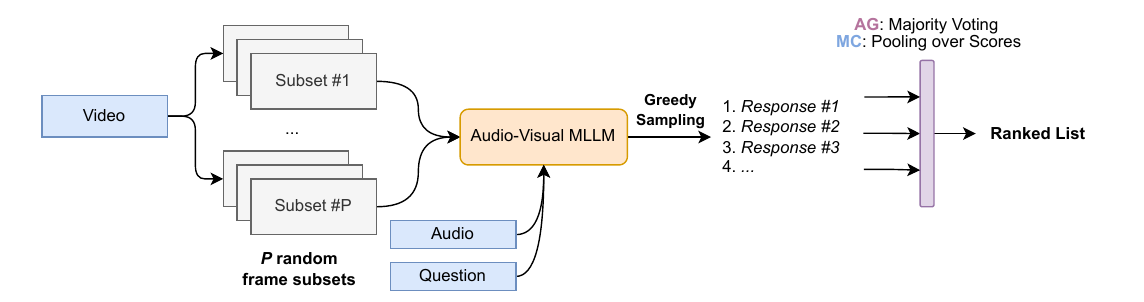}
\caption{Video Question Answering architecture.}
\label{fig:vqa_arch}
\end{figure}

The TRECVID 2025 Video Question Answering (VQA) track consists of the following two tasks: a) Answer Generation (AG), where the objective is to generate a ranked list of up to 10 natural language responses to a video question prompt and b) Multiple Choice (MC), where the objective is to rank 4 candidate responses to a video question prompt.
The evaluation dataset consists of 2,000 short videos that contain both audio and visual streams.

We address both tasks by employing a state-of-the-art audio-visual Multimodal LLM. We fine-tune the base model on the TRECVID-provided training dataset. During inference, we scale the multimodal understanding capabilities of the MLLM by aggregating multiple generations from different frame subsets of the video, as explained in detail in the next sections.
The complete architecture is shown in Figure~\ref{fig:vqa_arch}.
We present the official evaluations of the runs and additional experiments to validate our methodology.

\subsection{Approach}

\subsubsection{Base Model Selection and Fine-tuning}
\label{sec:vqa_finetune}

In order to address the requirements of the Video Question Answering tasks, we selected an omni-modal model, i.e. a model that is capable of processing both the visual and audio contents of the videos. Specifically, we used the Thinker component of Qwen2.5-Omni 7B \cite{xu2025qwen2} as our base model, while disabling the Talker component. We fine-tuned the Multimodal LLM using standard Supervised Fine-tuning (SFT) on the training dataset provided by the TRECVID organizers for this task\footnote{Training dataset can be found here: \url{https://www-nlpir.nist.gov/projects/tv2025/vqa.html}}. This training dataset consists of 500 videos with audio and visual streams, along with associated question-answer pairs. In particular, for fine-tuning we used only the correct answers provided for each question and ignored the wrong answers. We fine-tuned the model on a single RTX 4090 24GB GPU using LoRA \cite{hu2022lora} for 5 epochs with batch size 32 and learning rate 3e-5. Due to memory and compute constraints, we opted to sample 10 frames per video during training and up to 1024 visual tokens per prompt. Regarding the LoRA adapter, we set $r=16$ and $\alpha=64$.

\subsubsection{Frame Subset Ensembling}
\label{sec:vqa_self}

A well-known methodology for enhancing the capabilities of Large Language Models is self-consistency \cite{wang2022self}. The main idea is to generate a diverse set of outputs and aggregate them, typically via majority voting, thus reducing the probability of error due to a single erroneous generation. In the video domain, Suo et al. \cite{suo2025trial} proposed a similar idea addressing the Long Video Understanding task. Specifically, they proposed sampling different subsets of frames bin-wise, generating different predictions for each subset, and aggregating the generations with a self-reward that is a linear combination of three scoring functions. Video Parallel Scaling (VPS) \cite{chung2025video}, which is contemporaneous with our work, structures this process as an inference-time parallel generation task and aggregates the output per-token with a weighted average. Finally, VidCtx \cite{goulas2025vidctx} employs a similar frame-wise process, generating multiple predictions per video with an image MLLM that observes a single frame and relevant context, while aggregating the predictions across frames with max pooling.

Motivated by the above works, we adapt this paradigm to the open-ended audio-visual Video Question Answering task. In contrast to Suo et al. \cite{suo2025trial}, we employ a simple majority voting aggregation layer, without introducing external supervision from additional expensive models. Critically, we generate one LLM completion per frame subset using greedy sampling. Greedy sampling significantly reduces the exploration factor during auto-regressive generation, which enables us to utilize majority voting with open-ended free-form responses. Therefore, the response diversity is exclusively due to the sampling of different frames per generation. Furthermore, we do not use bin-wise frame sampling, considering the short duration of the videos.

We sample $P=10$ frame subsets randomly from the entire video, considering all frames. Each subset can have up to $N=2fps \times T sec$ frames, where $T$ is the duration of the video in seconds. However, due to the sampling process, the FPS of each subset is not constant throughout the entire duration. Furthermore, we combine the audio embeddings of the entire video with the sampled frames as-is. During inference, we set the maximum number of visual tokens to $16,384$ for each generation. Finally, we rank the generated responses based on the number of occurrences, i.e. the number of votes, of each response.

For the VQA Multiple-Choice (MC) task, we follow a similar approach as described above.
Specifically, we instruct the Multimodal LLM to generate the letter of the predicted answer (i.e., A, B, C, D). Following \cite{goulas2025vidctx} and \cite{suo2025trial}, we utilize a logit pooling layer instead of majority voting. This layer aggregates the log-probabilities (logits) assigned by the Multimodal LLM to the letters of the candidate choices across the responses. Specifically, we use mean pooling and we rank the responses according to the pooled probabilities. In addition, we experimented with ranking the candidate responses by estimating the confidence assigned by the model to each one, via the mean log probability of the response tokens. However, in our experimental setup, this method of assigning scores to each candidate response significantly underperformed the direct generation of the letter of the predicted answer, as shown in the next section.

\subsection{Submission - Answer Generation}

For the VQA Answer Generation task, we submitted the following 4 runs:
\begin{itemize}
\item certh.vqa.ag.run\_1: (Main run) Aggregation of $P=10$ predictions per question from different frame subsets of the video, generated with the fine-tuned model via greedy sampling.
\item certh.vqa.ag.run\_2: $P=1$ generation per question with the fine-tuned model.
\item certh.vqa.ag.run\_3: Same as Run 1, but with multinomial sampling instead of greedy sampling.
\item certh.vqa.ag.run\_4: Same as Run 1, removing any quote characters from the final responses as a post-processing step. Run 4 was submitted as a check to determine whether the evaluation platform parsed the submitted CSV files as intended.
\end{itemize}

\subsection{Submission - Multiple Choice}
For the VQA Multiple Choice task, we submitted the following 4 runs:
\begin{itemize}
\item certh.vqa.mc.run\_1: Base model without fine-tuning. Aggregation of $P=10$ predictions with mean pooling of the candidate logits.
\item certh.vqa.mc.run\_2: Fine-tuned model. Aggregation of $P=10$ predictions with mean pooling of the candidate logits.
\item certh.vqa.mc.run\_3: Base model without fine-tuning. Aggregation of $P=10$ predictions, where the answer is chosen by confidence estimation.
\item certh.vqa.mc.run\_4: Fine-tuned model. Aggregation of $P=10$ predictions, where the answer is chosen by confidence estimation.
\end{itemize}

\subsection{Experimental Results}

\subsubsection{Results for the Answer Generation Task}

\begin{figure}[tb]
\centering
\begin{tikzpicture}
\begin{axis}[
    ybar,
    width=0.95\textwidth,
    bar width=4pt,
    height=8cm,
    ylabel={Scores},
    xlabel={Runs},
    legend style={
        at={(0.97, 1.05)},
        anchor=north east,
        legend columns=-1
    },
    symbolic x coords={
        certh.vqa.ag.run\_2,
        certh.vqa.ag.run\_4,
        certh.vqa.ag.run\_1,
        videollama-2B,
        2B-without-prompt,
        videollama-2B-prompt,
        videollama-7B-prompt,
        certh.vqa.ag.run\_3,
        nut-kslab-2025,
        MARS-GenR\_1,
        ag\_nii\_uit,
        gemini\_flash,
        gemini\_flash\_2\_5,
        gemini\_flash\_25,
        TJUMIV3-corrected,
        TJUMIV6-corrected,
        TJUMIV4-corrected,
        gemini\_flash\_lite,
        TJUMIV5-corrected
    },
    xticklabels={
        \textbf{\color{red}certh.vqa.ag.run\_2},
        \textbf{\color{red}certh.vqa.ag.run\_4},
        \textbf{\color{red}certh.vqa.ag.run\_1},
        videollama-2B,
        2B-without-prompt,
        videollama-2B-prompt,
        videollama-7B-prompt,
        \textbf{\color{red}certh.vqa.ag.run\_3},
        nut-kslab-2025,
        MARS-GenR\_1,
        ag\_nii\_uit,
        gemini\_flash,
        gemini\_flash\_2\_5,
        gemini\_flash\_25,
        TJUMIV3-corrected,
        TJUMIV6-corrected,
        TJUMIV4-corrected,
        gemini\_flash\_lite,
        TJUMIV5-corrected
    },
    xticklabel style={
        rotate=45,
        anchor=north east,
        font=\small
    },
    xtick=data,
    ymin=0,
    ymax=100,
    ymajorgrids=true,
    yminorgrids=true,
    minor y tick num=1,
    ytick distance=10,
    minor grid style={dotted, gray!60},
    enlarge x limits=0.05,
    major x tick style={draw=none}
]
\addplot[fill=blue!60] coordinates {
(certh.vqa.ag.run\_2, 26.0)
(certh.vqa.ag.run\_1, 26.0)
(certh.vqa.ag.run\_4, 26.0)
(videollama-2B, 23.5)
(2B-without-prompt, 24.2)
(videollama-2B-prompt, 23.1)
(videollama-7B-prompt, 23.8)
(certh.vqa.ag.run\_3, 20.3)
(nut-kslab-2025, 22.6)
(MARS-GenR\_1, 21.9)
(ag\_nii\_uit, 17.3)
(gemini\_flash, 21.3)
(gemini\_flash\_2\_5, 21.3)
(gemini\_flash\_25, 21.3)
(TJUMIV3-corrected, 21)
(TJUMIV6-corrected, 21)
(TJUMIV4-corrected, 21)
(gemini\_flash\_lite, 18.9)
(TJUMIV5-corrected, 12)
};
\addplot[fill=red!60] coordinates {
(certh.vqa.ag.run\_2, 89.8)
(certh.vqa.ag.run\_1, 90.3)
(certh.vqa.ag.run\_4, 90.3)
(videollama-2B, 86.6)
(2B-without-prompt, 88.1)
(videollama-2B-prompt, 86.7)
(videollama-7B-prompt, 86.5)
(certh.vqa.ag.run\_3, 89.1)
(nut-kslab-2025, 89.3)
(MARS-GenR\_1, 88.9)
(ag\_nii\_uit, 88.7)
(gemini\_flash, 86.2)
(gemini\_flash\_2\_5, 86.2)
(gemini\_flash\_25, 86.2)
(TJUMIV3-corrected, 87)
(TJUMIV6-corrected, 87)
(TJUMIV4-corrected, 87)
(gemini\_flash\_lite, 85.6)
(TJUMIV5-corrected, 87)
};
\addplot[fill=yellow!60] coordinates {
(certh.vqa.ag.run\_2, 35.1)
(certh.vqa.ag.run\_4, 34.8)
(certh.vqa.ag.run\_1, 34.8)
(videollama-2B, 30.2)
(2B-without-prompt, 30.1)
(videollama-2B-prompt, 30.1)
(videollama-7B-prompt, 29.9)
(certh.vqa.ag.run\_3, 28.5)
(nut-kslab-2025, 28.4)
(MARS-GenR\_1, 28.2)
(ag\_nii\_uit, 27.0)
(gemini\_flash, 25.7)
(gemini\_flash\_2\_5, 25.7)
(gemini\_flash\_25, 25.7)
(TJUMIV3-corrected, 25)
(TJUMIV6-corrected, 25)
(TJUMIV4-corrected, 25)
(gemini\_flash\_lite, 21.6)
(TJUMIV5-corrected, 15)
};
\legend{METEOR, BERT, STS}
\end{axis}
\end{tikzpicture}
\caption{VQA Answer Generation (AG) ranking of all submitted runs in TRECVID 2025, sorted by STS score. Run IDs in red on the horizontal axis indicate our submitted runs.}
\label{fig:vqa_ag_ranking}
\end{figure}
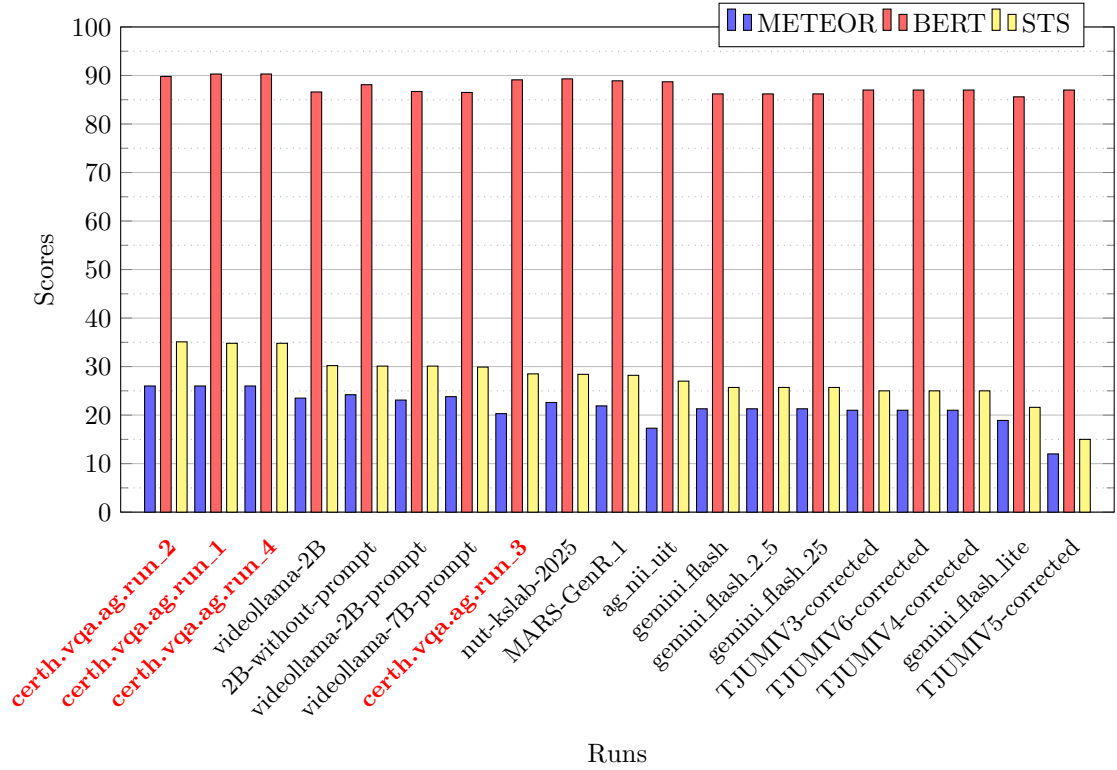

\begin{table}[htb]
\caption{Official results for the submitted runs in the VQA Answer Generation (AG) task. Best results are \textbf{emphasized}. The results of Run 4 are {\color{gray}de-emphasized} because the run is the same as Run 1, with all quote characters removed.}
\label{tab:vqa_ag_results}
\begin{center}
\begin{tabular}{l|ccccc}
\hline
\textbf{Run id:} & \textbf{METEOR} & \textbf{BERTScore} & \textbf{STS} \\
\hline
certh.vqa.ag.run\_1 & \textbf{0.260}* & \textbf{0.903}* & 0.348 & \\
\hline
certh.vqa.ag.run\_2 & \textbf{0.260}* & 0.898 & \textbf{0.351}* \\
\hline
certh.vqa.ag.run\_3 & 0.203 & 0.891 & 0.285 \\
\hline
\color{gray} certh.vqa.ag.run\_4 & \color{gray}0.260 & \color{gray}0.903 & \color{gray}0.348 \\
\hline
\end{tabular}
\end{center}
\footnotesize * We denote with an asterisk the runs that achieved the highest score among all participants in TRECVID 2025 for each metric.
\end{table}

\begin{table}[htb]
\caption{Results with top-1 selection for the VQA Answer Generation (AG) task. Best results are \textbf{emphasized}.}
\label{tab:vqa_ag_results_top1}
\begin{center}
\begin{tabular}{l|ccc}
\hline
\textbf{Run id:} & \textbf{METEOR} & \textbf{BERTScore} & \textbf{STS} \\ \hline
certh.vqa.ag.run\_1 + Top-1 & \textbf{0.265} & \textbf{0.905} & \textbf{0.356} \\ \hline
certh.vqa.ag.run\_2 + Top-1 & 0.260 & 0.898 & 0.351 \\ \hline
certh.vqa.ag.run\_3 + Top-1 & 0.215 & 0.897 & 0.304 \\ \hline
\end{tabular}
\end{center}
\end{table}

We present the official results of our submitted runs in Table~\ref{tab:vqa_ag_results}. The runs are evaluated based on the following 3 metrics, averaged across the responses: a) METEOR score, b) BERT semantic similarity score and c) Semantic Textual Similarity (STS) score.

Firstly, we observe that our main run (Run 1) significantly outperforms Run 3, which uses multinomial sampling instead of greedy sampling, due to the reduced randomness in the generated responses. Secondly, Run 1 and Run 4 have identical performance, since the only difference between the two runs is the removal of all quote characters from the final responses, as explained in the previous section. Therefore, we de-emphasize the results of Run 4 in the table. Third, the difference between our main run (Run 1) and the single-generation baseline (Run 2) is more subtle. However, as we explain in the next section, the two runs are not directly comparable, considering the different number of responses submitted per question between the two.

Figure~\ref{fig:vqa_ag_ranking} shows the performance of our submitted runs in the VQA AG track compared to all submitted runs from the other participating teams, sorted by STS score. We observe that we achieve the highest scores among all participants, across all 3 evaluation metrics.

\subsubsection{Discussion on Answer Generation Task and Additional Evaluation}

We observe that the aggregation of the metrics (METEOR, BERTScore, STS) across the ranked responses penalizes the methods that return more responses per question. Even if the response with the highest rank is correct, the presence of additional responses has a significant negative impact on the final score. Therefore, for a fair comparison, we present an additional evaluation of the runs, where we keep the highest ranked answer and discard the rest (i.e., employing top-1 selection). We present the results of this evaluation in Table~\ref{tab:vqa_ag_results_top1}. This process ensures that all runs have exactly one response. We calculate the new metrics on the basis of the provided per-response evaluation CSV files. We observe that under this setup, our main run (Run 1) significantly outperforms the baseline approach (Run 2) across all metrics, validating our methodology. In addition, we see that multinomial sampling (Run 3) continues to have a significant negative impact on performance, as explained above.

\subsubsection{Results for the Multiple Choice Task}

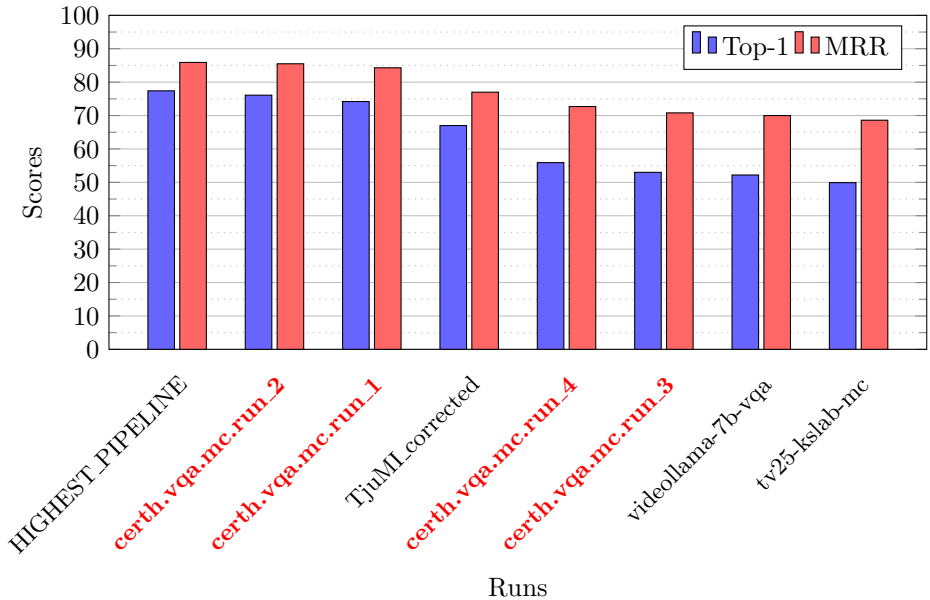
\begin{figure}[htb]
\centering
\begin{tikzpicture}
\begin{axis}[
    ybar,
    width=0.8\textwidth,
    bar width=10pt,
    height=6cm,
    ylabel={Scores},
    xlabel={Runs},
    legend style={
        at={(0.97, 0.97)},
        anchor=north east,
        legend columns=-1
    },
    symbolic x coords={
        HIGHEST\_PIPELINE,
        certh.vqa.mc.run\_2,
        certh.vqa.mc.run\_1,
        TjuMI\_corrected,
        certh.vqa.mc.run\_4,
        certh.vqa.mc.run\_3,
        videollama-7b-vqa,
        tv25-kslab-mc
    },
    xticklabels={
        HIGHEST\_PIPELINE,
        \textbf{\color{red}certh.vqa.mc.run\_2},
        \textbf{\color{red}certh.vqa.mc.run\_1},
        TjuMI\_corrected,
        \textbf{\color{red}certh.vqa.mc.run\_4},
        \textbf{\color{red}certh.vqa.mc.run\_3},
        videollama-7b-vqa,
        tv25-kslab-mc
    },
    xticklabel style={
        rotate=45,
        anchor=north east,
        font=\small
    },
    xtick=data,
    ymin=0,
    ymax=100,
    ymajorgrids=true,
    yminorgrids=true,
    ytick distance=10,
    minor y tick num=1,
    minor grid style={dotted, gray!60},
    enlarge x limits=0.1,
    major x tick style={draw=none}
]
\addplot[fill=blue!60] coordinates {
(HIGHEST\_PIPELINE, 77.4)
(certh.vqa.mc.run\_2, 76.1)
(certh.vqa.mc.run\_1, 74.2)
(TjuMI\_corrected, 67)
(certh.vqa.mc.run\_4, 55.9)
(certh.vqa.mc.run\_3, 53.0)
(videollama-7b-vqa, 52.2)
(tv25-kslab-mc, 49.9)
};
\addplot[fill=red!60] coordinates {
(HIGHEST\_PIPELINE, 85.9)
(certh.vqa.mc.run\_2, 85.5)
(certh.vqa.mc.run\_1, 84.3)
(TjuMI\_corrected, 77)
(certh.vqa.mc.run\_4, 72.7)
(certh.vqa.mc.run\_3, 70.8)
(videollama-7b-vqa, 70)
(tv25-kslab-mc, 68.6)
};
\legend{Top-1, MRR}
\end{axis}
\end{tikzpicture}
\caption{VQA Multiple Choice (MC) track ranking of all submitted runs in TRECVID 25 sorted by top-1 accuracy. Run IDs in red on the horizontal axis indicate our submitted runs.}
\label{fig:vqa_mc_ranking}
\end{figure}

\begin{table}[htb]
\caption{Official results for the VQA Multiple Choice (MC) task. Best results are \textbf{emphasized}. SFT refers to Supervised Fine-tuning.}
\label{tab:vqa_mc_results}
\begin{center}
\begin{tabular}{c|c|cc}
\hline
\textbf{Run id:} & \textbf{Variant} & \textbf{Top-1 (\%)} & \textbf{MRR} \\ \hline
certh.vqa.mc.run\_1 & zero-shot & 74.2 & 0.843 \\ \hline
certh.vqa.mc.run\_2 & SFT & \textbf{76.1} & \textbf{0.855} \\ \hline
certh.vqa.mc.run\_3 & zero-shot & 53.0 & 0.708 \\ \hline
certh.vqa.mc.run\_4 & SFT & 55.9 & 0.727 \\ \hline
\end{tabular}
\end{center}
\end{table}

\begin{table}[htb]
\caption{Ablation study for the VQA Multiple Choice (MC) task. Best results are \textbf{emphasized}.}
\label{tab:vqa_mc_ablation}
\begin{center}
\begin{tabular}{l|cc}
\hline
\textbf{Method} & \textbf{Top-1 (\%)} & \textbf{MRR} \\ \hline
Base model & 73.2 & 0.838 \\
+ multiple generations & 74.2 & 0.843 \\ \hline
Fine-tuned model & 74.4 & 0.845 \\
+ multiple generations & \textbf{76.1} & \textbf{0.855} \\ \hline
\end{tabular}
\end{center}
\end{table}

In Table~\ref{tab:vqa_mc_results}, we present the official results of the submitted runs in the Video Question Answering Multiple Choice (MC) task.
The metrics that were used to evaluate the runs are a) the top-1 accuracy and b) the mean reciprocal rank.
Firstly, we observe that both for direct generation and confidence estimation, our fine-tuned model outperforms the base Qwen2.5-Omni model, even though it was trained on open-ended question-answer pairs.
Secondly, we observe that the direct generation method significantly outperforms the confidence estimation method.

Figure~\ref{fig:vqa_mc_ranking} shows the performance of our submitted runs in the VQA MC track compared to all submitted runs from the other participating teams, sorted by top-1 accuracy. We observe that our approach achieves the second highest score across all participants.

To validate the performance of the scaling process, we performed additional ablation experiments, which are presented in Table~\ref{tab:vqa_mc_ablation}.
We evaluate the new runs based on the provided ground truth responses from the organizers.
Based on these experiments, we conclude that the inference-time technique provides a clear performance gain in both setups.

\textbf{Note about the evaluation.}
We observed that a small number of multiple-choice questions and candidate answers in the provided evaluation dataset contained incorrectly escaped quotes (i.e., a single quote character instead of the expected double quote). We manually escaped the quotes for those entries prior to submitting the official results. However, according to the provided diagnostic CSV files, these responses were all marked as incorrect by the automatic evaluation system due to incorrect parsing. This issue affects 0.7\% of the questions. In order to present a fair comparison of the ablation experiments with the official runs that we submitted, we, too, mark those answers as incorrect in the ablation study. 

\section{Conclusions}

In conclusion, in the AVS track, we presented a two-stage AVS framework that leverages foundation models for both retrieval and reranking. We improve the performance of our pipeline through MLLM-based reranking; this revises the initial retrieval results by improving the semantic alignment between the query and the top-ranked shots, consistently producing more accurate and contextually relevant final rankings. 

Regarding the VQA track, through the combination of fine-tuning and inference-time scaling with the aggregation of multiple predictions per question, generated from different frame subsets, we achieve the highest score across the employed 3 metrics in the Answer Generation task, and the second-highest score in the Multiple Choice task. Furthermore, we show through additional experiments the effectiveness of each component of our framework and we present additional evaluations of the submitted runs, for a more direct comparison.

\section{Acknowledgements}
This work was supported by the EU’s Horizon Europe programme under grant agreements 101070190 AI4Trust and 101070109 TransMIXR.


\end{document}